# Classifying Dominant Temporal Orientation without Pretrained Text Embeddings: A Novel Morphosyntactic Inventory Vector Approach

**Jonathan Cleveland[1] and Peter S. Bearman[1]**
[1]Interdisciplinary Center for Innovative Theory and Empirics, Columbia University, New York, NY 10115**,** Department of Sociology, Columbia University, New York, NY 100271
To whom correspondence may be addressed: jc2457columbia.edu or psb17@columbia.edu.

**Abstract**
Computational methods have consistently struggled to determine the dominant temporal orientation of a sentence. This difficulty is especially pronounced when a sentence contains multiple embedded clauses with competing tense and aspectual information. To address this difficulty, we propose an alternative approach for identifying a sentence's past, present, or future global reference interval. Our method does not use any form of pretrained embeddings. We rather encode sentences using fixed-length inventory vectors that are comprised of part-of-speech counts, dependency relation counts, and explicit futurate pattern counts. We term this inventory vector of a sentence a "Morphosyntacton". The method does not use any padding, sequence models, or large language models. Evaluation on 1,799 syntactically complex English sentences, annotated as past, present or future, shows balanced and high accuracy multiclass classification, achieving an overall multi-class accuracy of 92%.

## I. INTRODUCTION

Programmatic methods to extract the temporal meaning of English sentences possess a long-standing history of being a 'hard problem.' In computational linguistics, this difficulty occurs broadly among studies that focus on extracting the temporal references of events, temporal relations between events and time expressions (Chambers, Wang, and Jurafsky 2007; Choubey and Huang 2017; Huang et al. 2016, 2023; Xu et al. 2022). Outside of computational linguistics, it is primarily in computational psychology where this difficulty manifests among the more restricted efforts to extract an actor's temporal orientation from textual sources (Filannino and Nenadic 2014; Kamila et al. 2018; Park et al. 2017; Schwartz et al. 2015; Walsh and Busby Grant 2018)**.** In these approaches, textual sources are argued to be reflective of the temporal orientations present in the composer's actual thought process. Usually, these approaches seek to quantify the presence of a single-dominant temporal focus by summing the individual temporal intents of subunits of a text into one dominant net temporal focus.

A problem with many of the above studies is their suboptimal overall accuracy scores.

For example, in the more granular, event focused task of computational linguistics, Huang's accuracy scores on lengthy complex newspaper articles achieved: 87% accuracy on determining whether a previous event occurred; 51% accuracy on determining whether a present event is ongoing; 57% accuracy on references to future events (i.e., future planned or future possible) (Huang et al. 2016). Whereas in the more restrictive focus of computational psychology, Kamila et al. using simple tweets achieved accuracy scores of 81.75% on past, 61.71% on present, 73.07% on future references (Kamila et al. 2018). In addition, Schwartz et al., using Facebook posts, achieved 71% on past, 79 % on present, and 53% on future (Schwartz et al. 2015). Also, Park et al., using Facebook data, appears to roughly achieve overall 72 % accuracy (Park et al. 2017b). Similarly, Filannino and Nenadic, achieved roughly overall 72.33% on predicting the temporal orientation of search engine queries (Filannino and Nenadic 2014). Overall, most of these types of studies usually show higher accuracy scores on past references in comparison to present or future references. In addition, all of these types of studies rely on some form of pretrained text embedders, whether legacy forms such as Word-to-Vec or the more contemporary state-of-the-art transformers.

The comparatively low accuracy scores above can also be found when using Large Language Models (LLM). Alsayyah and Batista-Navarro argued that use of an LLM such as ChatGPT for temporal relation extraction "is not comparable in relation to that of supervised models (Alsayyah and Batista-Navarro 2023:8)." Using the TIMELINE test set to extract temporal relations between events, the authors used a zero-shot prompt and recorded precision, recall, and F1 scores of 31.11%, 35.67% and 33.24%, respectively (Alsayyah and Batista-Navarro 2023:8). Similarly, Yuan et al. used a zero-shot temporal relation prompt on Chat GPT to extract temporal relations and found its performance can be up to 30% worse than a supervised model based on F1-scores (Yuan, Xie, and Ananiadou 2023).

Another significant issue in many of the aforementioned studies is their tendency to be unable overall to classify accurately sentences which contain what Klein broadly categorized as a poly-temporal construction (Klein 1994:36). From a linguistics perspective, a machine classifier can especially struggle with a sentence that contains multiple embedded clauses of differing aspect morphologies (Smith 1991). In fact, this difficulty may explain why many studies in this problem domain tend to focus on datasets that contain extremely short snippets of text found in collections of tweets or Facebook posts. Despite the use of transformers, lengthy sentences appear as a major obstacle in machine-based efforts to assign a sentence with a single global-dominant temporal intent.

## 1.0 A New Approach to Encoding Temporal Dimensions of Sentences

In our study, we offer an alternative to the above methods used for extracting the dominant temporal meaning of lengthy sentences that contain significant clausal complexity. Our method is based on *Morphosyntactic Inventory Vectors*. We term this unified, representation of a sentence's morphosyntactic organization a "Morphosyntacton." This approach does not use any form of pretrained text embedders. We alternatively obtain vector representations of our sentences by using NLP techniques to inventory each individual element of a sentence for 49 part of speech tags and 45 dependency tags. In addition, we inventory each sentence using a series of functions to pick up on 7 different futurate references. Our technique vectorizes the counts of these sets of tags and the counts generated from our futurate functions together into a single data frame. No padding techniques to equalize sentence length are used in our method. No large language models are used.

### 1.1 Morphosyntactic Inventory Vectors Yield Improved Accuracy

Our model was trained on a dataset of 1,799 sentences characterized by significant syntactic/clausal complexity (average words/sentence = 16.06, average tree depth = 5.93; average T-units per sentence = 1.29). Sentences contained references to complex eventive structures. The model demonstrated strong performance across three broad temporal categories. Specifically, harmonic F1 scores for past, present, and future were 0.94, 0.90, and 0.91, respectively. The overall multi-class accuracy was 0.92.

### 1.2 Course of the Paper

In what follows, we first background the problem domain of clausal structure and asymmetric aspect morphology in machine-based temporal classification. We suggest this problem serves as one of the primary stumbling blocks for the lower accuracy scores seen in some machine-based temporal classification studies. After this, we offer a critique of the ubiquitous use of tweets and Facebook data for machine-based temporal extraction, raising the question of whether this data type presents unique problems for temporal extraction. We then give a detailed description of our dataset and our annotation policy. This policy offers the reader specific exemplars in granular detail of sentences used as models for what constitutes a past, present or future oriented sentence.

Our approach above differs from many of the above studies, as we neither rely on or employ inter-rater agreement scores. In alignment with Alsayyah and Batista-Navarro's efforts to highlight the inherent difficulty of temporal-based annotation (Alsayyah and Batista-Navarro 2023:1), we suggest that a lack of a sufficiently specific annotation policy is another stumbling block involved in the lower accuracy scores seen in some studies. After this, we then describe in detail our above-mentioned technique for extracting the dominant temporal meaning of a sentence and the resulting accuracy scores this technique yielded on our test set of sentences - comparing performance without and with pretrained embeddings attached to our vector representation.

## 2.0 Problem Domain: Clausal Subordination & Asymmetric Aspect Morphology

A machine classifier can struggle with the types of contradictions that occur between clausal subordination and dominant aspect morphology. For a human classifier, the capacity to differentiate clause structures that reflect the primary vs. subordinate temporal reference categories serves as a defining feature of linguistic competence. However, even a human classifier can find this relationship challenging. This challenge can be especially pronounced when dealing with finite clauses (Biber and Gray 2010) or non-finite clauses that contain their own temporal adjunct (Saurí et al. 2004:40–42) . For example, in the following sentence, we see three different aspect morphologies corresponding to three different clauses

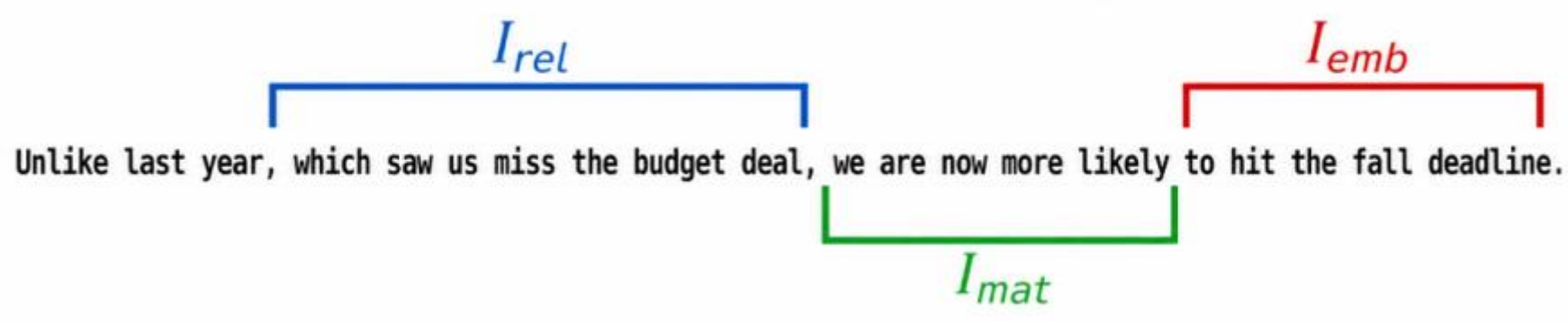

In the above, the relative clause, ($I_{rel}$) has an event interval that is occurring before the present now (R) ($I_{rel} < R$). Grammatically this past intention is achieved by a simple past perfective (PST.PFV) which suggests a completed, bounded past event. The relative pronoun 'which' refers back to the head noun 'last year', which functions as a temporal adjunct. However, this past focus changes in the matrix clause ($I_{mat}$). Here $I_{mat}$ is at the moment of utterance R ("now") which makes the clause present oriented ($R \in I_{mat}$). Grammatically, this present emphasis occurs as a result of a present tense 'are' which by default locates the **state** at the speech time. An indexical adverb 'now' places the utterance point at the deictic center (R). The epistemic modal 'more likely' gets evaluated at the reference time rather than shifted into the past or future. However, this focus on the temporal present changes in the embedded clause ($I_{emb}$) where $I_{emb} >$ R. Here "to hit the deadline" is located after that reference time and therefore the clause is strongly future bounded (Telic bounded). Grammatically, this future focus is achieved through a bare-infinitive that has a futurate function.

A machine-based classifier can have significant difficulty mirroring how a linguistically competent speaker can subordinate the multiple competing aspect-clauses present above into what Smith termed a global reference interval ($I_{dom}$) (Smith 1991:48) **.** Here ($I_{dom}$) is the maximum of its component intervals, where the max function returns the interval that is furthest to the right on the timeline

$$\textbf{Let} \quad I_{dom} = \max\{I_{rel}, I_{mat}, I_{emb}\}$$

$$\textbf{Given} \quad I_{emb} > I_{rel} \quad \wedge \quad I_{emb} > I_{mat}$$

$$\therefore \quad I_{dom} = I_{emb}$$

This basic illustration suggests that the more clausal structures a sentence contains, the more likely it is to exhibit conflicting aspect morphology. This follows proportionally, since measures of clausal density in a sentence can be tightly coupled with measures of the mean length of T units (MLTU) (Biber 1988:127–29; Hunt 1965; O'Donnell, Griffin, and Norris 1967; Wolfe-Quintero, Inagaki, and Kim 1998). We can formalize this relationship symbolically with the following approximate representation:

$$C_i = \beta \cdot MLTU_i$$

$C_i$ is the (expected) number of clauses in sentence i, where $MLTU_i$ is the mean length of T-units in that sentence and β is a proportionality constant representing average clause density per T-unit. As the mean length of MLTU increases in a sentence ($MLTU_i$) the number of clauses tends to increase ($C_i$). This rise in structural complexity leads to a higher probability of aspectual contradictions. In turn, the implication is that lengthy sentences introduce more opportunities for conflicts between aspect markings in subordinate clauses and a matrix clause.

**2.1 Aspect Morphology in Tweets and Facebook Posts: Solution or Added Problem?**

From a machine learning perspective, the consequence of what we may term the 'aspect-clausal schism' may generate an empirical necessity to focus on short forms of text, such as tweets or Facebook posts. It should be kept in mind that the average tweet in 2012 was reported to be 28 characters long (SMK: Social Media Knowledge 2012). The average Facebook post in 2013 was reported to be 155 characters long (Gessler 2016). Syntactical studies of tweets have found a very strong presence of elliptical clauses (Crystal 2011:45). Not surprisingly, Crystal's study found that conditional clauses and relative clauses were each individually under twenty percent of the sampled corpora (Crystal 2011:8). From these occurrences, tweets and related Facebook posts might appear to be more tractable for extracting temporal references from text. There are, however, unique problems that are likely to occur when training a classifier on texts with high rates of elliptical clauses.

Economic tweets or Facebook posts are known to be syntactically minimal (Thurlow and Mroczek 2011) . Subjects and auxiliaries can be omitted, creating a highly fragmented domain for deciphering meaning (Boyd, Golder, and Lotan 2010; Efron and Winget 2010; Zappavigna 2015). Zappavigna's analysis of the 100-million-word Twitter dataset (HERMES) emphasized that tweets often omit explicit temporal connectives (Zappavigna 2015:27–49). They can rather rely on pragmatic inference or contextual metadata to decipher the temporal meaning of events (Zappavigna 2015:151–69). The lower accuracy scores in the above studies that relied on tweets and Facebook data suggest that this type of lexico-syntactical impoverishment could be a significant culprit. Stated another way, substantial lexical and syntactical references to time are likely more than necessary when trying to use a machine classifier to objectify the nebulous and contradictory realm of time reflections in language. An example of this criticism in relation to tweets is presented in the annotation criticism section below.

### 3.0 Annotation Criticism

Many of the aforementioned studies tend to have inter-rater-agreement scores in the eighty percent range. For example, Schwartz et al. reported an inter-rater agreement coefficient of 0.85 (Schwartz et al. 2015:413). Park et al. had an inter-rater agreement of 0.85 (Park et al. 2017:272) and Huang et al. had 0.84 (Cohen's K) (Huang et al. 2016). Kamila et al. had a multi-rater kappa agreement of 0.82 (Kamila et al. 2018:668). In our view, a 15% discordance rate raises concerns about the consistency of the human classifications, which are supposed to represent a 'ground truth'. In general, it should be appreciated that the manual annotation of temporal references derived from textual sources is not a trivial task. As Verhagen et al. argued in relation to the TempEval framework, "annotation of temporal relations is not an easy task for humans due to rampant temporal vagueness in natural language (Verhagen et al. 2007:76)."

Some of the low classification accuracy scores cited above - particularly on present and future references - reinforce the above concerns, since there is the intuitive expectation that inconsistent classification of the training set will degrade classifier accuracy. A possible reason for rater discordance can derive from the annotation policies given to raters. Rater discordance may specifically arise from annotation policies that lack specificity. This issue may be especially important when dealing with the types of registers present in tweets or Facebook posts. This follows from studies that suggest these forms of media are more reflective of how people actually talk (Tagg 2015; Thurlow and Mroczek 2011). In the specific case of tweets, Wikström described them as *"I tweet like I talk"* (Wikström 2016). The idea here is that Twitter communication has blurred the types of boundaries that exist between speech and writing.

Schwartz et al. provide an example of the impact of inter-rater discordance and the abbreviated structure typical of tweets and Facebook posts (Schwartz et al. 2015). This study focused on extracting an actor's temporal orientation. Below are a majority rule annotator table and caption from the study of Schwartz et al. The table contains, what Schwartz et al. term three 'easy' samples, with no discordance and three 'difficult' samples that generate discordance.

**Table 1 (from Schwartz et al.)**

| Status | R1 | R2 | R3 | Maj |
|---|---|---|---|---|
| *:) today was actually pretty good* | pa | pa | pa | pa |
| *is listening to The Sad Cafe by The Eagles!* | pr | pr | pr | pr |
| *considering checking out base jumping and parkour some time in the future XP* | fu | fu | fu | fu |
| *I just watched Oprah and am posting what it was about.* | pa | pr | pa | pa |
| *really wanted a snow day, but probably not going to get one tomorrow. now homework.* | pr | fu | fu | fu |
| *Another day of great restraint.* | pa | pa | pr | pa |

Table 1: Examples of statuses annotated for temporal classes: past (pa), present/none (pr), and future (fu). R1, R2, R3: judgements from each rater; Maj: choice from majority voting. The bottom three examples illustrate difficult cases.

In the above table, if one focuses on the 'great restraint' example there is an absence of a verb or other temporal specifiers. Any attempt to classify this sample is, in our view, aberrant, since one is forced to *guess* if the speaker meant "Today was another day of great restraint" or "Today is another day of great restraint." or "Today will be another day of great restraint." We suggest that a preferred annotation policy would be the deletion of all snippets without verbs or other temporal specifiers. This, we suggest, is a prototypic problem that occurs with these forms of media in relation to annotating the dominant temporal intent.

For raters, the other representative criticism above is the importance of pre-defining what dominant temporal intent means. When you are dealing with the conflicting opacity of tweets and Facebook posts, this may very well be a necessity. For example, if we examine the following sentence from above in relation to Hans Reichenbach tripartite model — Event Time (E), Reference Time (R), and Speech Time (S) (Reichenbach 1947).

*I just watched Oprah and am posting what it was about.*

In the main clause (I just watched Oprah) the Event time and Reference time are both referring to the Recent Past. Both are also occurring before the speech time of the present (E = R < S). Whereas in the verb phrase (and am posting) the Event time, Reference Time and Speech Time are all referring to the present (E = R = S). In the embedded noun clause "what it was about," both the Event time and the Reference time are in the past, occurring before the current Speech time (E = R < S). The basic point is that no single clause temporally dominates (i.e. no $I_{dom}$); instead, all clauses tend to hold equal temporal status. In this sense, one can argue there is no real hierarchy; each clause stands on equal footing, without one subordinating the others.

**3.1 Key Takeaway Regarding Annotation**

Broadly speaking, the anchor point that a rater chooses in opaque and/or multi-clausal sentences with significant tree depth may be influenced by the temporal perspective of the annotator. A human language expert engaged in the annotation of textual data might very well focus on the event time. While a discourse analyst or social media researcher, by contrast, might prioritize Reference Time — focusing on how the speaker situates each event relative to the other events**,** rather than on when the events actually occurred. We suggest that one should not assume that linguistic competence with majority rule alone will resolve such differences — particularly when raters are asked to annotate sentences in isolation without access to the broader discourse context.

To address the above types of problems, we suggest that raters need to be given exemplars. By 'exemplars', we do not mean that raters should receive explicit instruction on the types of distinctions present between broad descriptions of event time and reference time. In practice, maintaining awareness of these distinctions can be quite difficult, as they are often challenging to disambiguate (Michaelis 2006)**.** But we do suggest exemplars should be used in relation to the purpose of the annotation, as there may be multiple legitimate ways to assign temporal categories to the same sentence. In addition, exemplars may help to deal with the types of temporal opacity that can occur with stilted tweets and complex multiclausal sentences. Below, in our dataset-creation section, we offer specific exemplars that were used in our annotation policy to address these types of issues.

## 4.0 Dataset Composition

Our dataset contains 1799 nearly balanced sentences: 654 Past sentences, 521 Present sentences, 624 Future sentences. No preprocessing steps associated with stop-word removal or forms of stemming or lemmatization were used. Our collection of sentences comes from two different sources: The State of the Union Corpus (SOU) (State of the Union Corpus (1790 - 2018) n.d.) and from an algorithm used for the programmatic generation of sentences (**see 4.1**).

In the former, sentences for the SOU contain transcribed speeches. These speeches have been given to Congress by a sitting president ever since Washington's 1790 first SOU speech. A SOU has been given every year with the exception of 1933. Washington and Jefferson are reported to have given the speech in person to Congress. However, this in-person performance was subsequently avoided until Woodrow Wilson took office in 1913. It was during this time that the SOU transformed from essentially being a record of activities of the executive department into a plan for the President's upcoming legislative agenda. The advent of radio and particularly television in the 1960's brought SOUs into the living rooms of Americans (Rule, Cointet, and Bearman 2015). Below are two representative examples of the types of SOU sentences present in our dataset:

**Table 2**: **Examples of SOU Sentences**

| I applaud the bipartisan support this Congress provided last year for our initiatives to help Russia, Ukraine, and the other states through their epic transformations. |
| --- |

Even with our tax cuts, taxes as a fraction of gross national product remain about the same as they were in 1970.

From a temporal semantic perspective, these types of sentences can be expected to be challenging for a classifier that is tasked with assigning one dominant temporal orientation for each sentence. In alignment with our analysis above, we see complex clauses (infinitival purpose clause, restrictive relative clause, matrix clause, comparative clause) with multiple contrasting aspect morphologies. However, unlike tweets or Facebook data, our sentences are not grammatically truncated; in fact, they are quite the opposite. These types of rich eventive sentences make up roughly 80 percent of our dataset. Detailed corpus linguistics of the complete dataset is presented below in Section 4.3.

### 4.1 Programmatic Generation of Sentences

This second component of our dataset is composed of programmatically generated sentences. These sentences derived from our training efforts. We noticed our classifier was making consistent types of temporal classification errors. To deal with this, we used a template to programmatically generate sentences of specific temporal classes. These sentences make up roughly 19 % of our dataset. We call this sentence generation process, "the Dot-Format Sentence Generator" based on Python code jargon.

### 4.2 Dot-Format Sentence Generator

The Dot-Format Generator is written in Python and uses a sentence template that contains a series of variables to vary the content of the sentence. This variation can provide nuanced sentences of different temporal orientations at scale. For example, below you see Python code for a sentence template, which can iteratively generate future-oriented sentences. The sentence below serves as an exemplar, where the curly brackets serve as *placeholders* for elements from a list of futurate temporal verbs (first position) and future temporal adjuncts in the second position (See Appendix A.17 and A.18, for the past and present sentence templates and sample sentences generated from those templates.)

**Figure 1: Futurate Sentence Template**

**fu_template**
**I { } go to Paris { }.**
**futurate_vbs**
**[will, might, ought to, plan to, expect to, hope to]**
**fu_adjuncts**
**[next week, next month, next year, tomorrow, in the future, after I graduate, sometime soon, before I retire, later]**

The above set of Python lists above can be used to programmatically generate novel sentences. Below are some examples:

**Table 3: Future Oriented Sentences Generated from a Future Sentence Template**

| Subject | | Verb | Temporal Adjunct |
|---|---|---|---|
| I | will | go to Paris | next week |
| I | will | go to Paris | next month |
| I | might | go to Paris | next week |
| I | might | go to Paris | next month |
| I | ought to | go to Paris | next week |
| I | ought to | go to Paris | next month |

### 4.3 Metrics of Complete Dataset.

Prior to evaluation, our dataset corpus was profiled using sentence-level lexical and syntactic complexity measures. Sentences were tokenized and parsed with spaCy. We used Python scripts to compute structural length (words, characters) lexical variability (Shannon entropy), and syntactic complexity from dependency parses (clause counts, dependency arcs, subordination-coordination ratio, prepositions, subjects, passive constructions, T-units, and tree depth). These measures characterize the dataset's linguistic heterogeneity independent of modeling (See Appendix for complete descriptions of equations).

**Figure 2: Summary of Corpus Metrics**

The lexical diversity of our dataset is suggested by an entropy number of 9.49. This moderately high level of entropy indicates there is greater word variability and less repetition in our dataset vocabulary (Juola 1998; McCarthy and Jarvis 2007). This finding makes sense in regards to the level of diverse communicative contexts present in a political corpus like the SOU. This number also suggests that the programmatic generation of sentences in the second part of our dataset does not contribute to fostering lexical homogeneity.

The syntactical diversity of our corpus is suggested by an average clause/sentence score of 2.67. This number reflects a moderately high number of embedded clauses, and thereby significant subordination. Higher numbers of subordination can be seen to reflect significant syntactic complexity (Liu and Hu 2021). This likelihood is further strengthened by our high average dependencies per sentence score of 18.15 (Liu 2008) (Prévost, Reitter, and Nenkova 2022). In addition, the elevated Subordination/Coordination Ratio score of 1.98 reflects a preference for hierarchical structuring.

The elevated Subordination/Coordination Ratio suggests a higher level of syntactic complexity, as subordinate constructions require more cognitive planning and grammatical precision than coordination (Biber, Gray, and Poonpon 2011). According to Biber, this kind of syntactic pattern is a hallmark of formal written registers - particularly academic, legal, and policy-oriented prose. This score dovetails with our Average Tree Depth of 5.93, which suggests richly embedded syntactic structures. On average, almost 6 steps are taken from the main verb down to the most deeply embedded word. Greater depth and longer links reflect high rates of syntactical complexity (Chen, Deng, and Liu 2021; Hudson 1995; Liu 2008). Based on all of these scores, our corpus appears to possess significant syntactical diversity.

### 5.0 Annotation Policy

Our annotation policy focused on standardizing what constitutes a past, present or future oriented sentence. This process relied on generating a series of sentence exemplars. By 'exemplars', we mean a standardized agreement regarding what types of sentences constitute a past, present or future oriented sentence. This process was carried out by the investigators of this study. After this was achieved, the same investigators working together subsequently used these exemplars to annotate 1799 sentences. It is possible that another investigator might claim that our annotations of specific sentences are in some way biased. However, this presumption of bias is explicitly presented in our approach through the use of temporal sentence exemplars that are offered to the reader. In addition, any potential bias is explicitly represented in our training set, which can be modified and tested experimentally (See appendix, for the complete annotation policies and exemplars for future, past, and present categories, including near versus distant future and conflicting temporal references.)

## II. METHOD DEVELOPMENT

### 6.0 Stage One: The Problem of Dictionary Methods

Our method extracts the dominant temporal reference of a sentence by combining a vectorized inventory of morphosyntactic information derived from part-of-speech (POS) tags and syntactic dependency relations. In addition, we apply a set of functions that inventory futurate patterns, which are likewise encoded as vector features.

Our vectorized approach differs from the well-known limitations of pure dictionary-based count approaches for identifying the dominant temporal reference of a sentence, such as present in the Linguistic Enquiry Word Count (Boyd n.d.:12). Even in domains where the linkage between tense morphology and dominant temporal interpretation is comparatively tight, a dictionary-based inventory of surface-level POS counts for a dominant temporal reference can be misleading. For example, consider the following sets of simple past-marked sentences as they would be treated by a universal POS-based inventory.

**Table 4: We had better buy a car.**

| TOKEN | POS TAGS | COUNT |
|---|---|---|
| We | PRON | 1 |
| had | VBD | 1 |
| better | ADV | 1 |
| buy | VB | 1 |
| a | DET | 1 |
| car | NOUN | 1 |

Although the verb *had* is tagged as VBD and would be treated as a past-tense feature in a POS-based count representation, it does not function as a temporal anchor for past time. Instead, the construction *had better + VP* encodes modal obligation with a futurate interpretation. As a result, any approach that relies on counts of past-tense verbs to infer a global dominant temporal reference would misclassify this sentence. A similar issue arises in subjunctive constructions:

**Table 5: If we were to buy a car…**

| Token | POS | Count |
|---|---|---|
| If | IN | 1 |
| we | PRP | 1 |

| Token | POS | Count |
|---|---|---|
| were | VBD | 1 |
| to | TO | 1 |
| buy | VB | 1 |
| a | DT | 1 |
| car | NN | 1 |

Here, *were* is morphologically past but serves as a marker of subjunctive mood, introducing a hypothetical condition rather than locating an event in past time. Once again, a dictionary-based strategy that equates VBD with a global, dominant past temporal reference would fail to recover the dominant temporal reference.

The important point is that although surface tense morphology may contradict temporal interpretation, systematic patterns may encode temporal meaning at deeper grammatical levels. These patterns may be represented in POS tags and other indexical tags, such as dependency relations, when these features are considered in co-occurrence rather than in isolation. However, the use of sequences of interrelated tags requires an encoding that makes them accessible to a machine learning classifier. Our solution to this problem is to use what we call a **Morphosyntactic Inventory Vector Space.** This space is comprised of a Part of Speech Inventory Vector Space, a Dependency Relation Inventory Vector Space and a Futurate Inventory Vector Space.

**6.1 Stage Two: Part of Speech Inventory Vector Space.**

Machine learning classifiers of textual data are dependent on vectors. It is through numeric vectorization of text that a classifier can find meaning-based patterns. The first problem concerning the vectorizing of POS tags is that vectors in a vector space must all be of the same dimensionality. Simply stated: sentences of different lengths generate accompanying lists of POS tags of different lengths.

To deal with the above problem, we used the complete set of fine-grained POS tags for words and punctuation marks available in the spaCy 3.5.0 software. For each sentence, 49 POS tags were inventoried using spaCy's fine grained POS tags (Honnibal et al. 2020). This approach allows sentences of any length to be considered and gives a 49-dimensional space, with one dimension for each POS tag. At each element in our POS-inventory vector, we set the value to the number of times that the corresponding POS tag occurs in the sentence. This means that if a POS tag is not found in a sentence, it is counted as a zero in the corresponding element of the vector. Each POS-Inventory vector, therefore, specifies the POS tags that describes the sentence, as those that do not.

The above process can be represented as follows: consider a sentence ***s*** consisting of an ordered list of tokens, such that $w_1$ is the first token and $w_n$ is the last: $s = [w_1, w_2, \dots, w_n]$. Let **u(*s*)** denote the POS-count vector of the sentence *s:* $u(s) = [u_1(s), \dots, u_{49}(s)]$. Let $\{p_1, p_2, \dots, p_{49}\}$ be the fixed, ordered inventory of all 49 spaCy 3.5.0 defined fine-grained POS tags. Using the 49 spaCy functions, we get:

$$\mathbf{u}(s) = \left[ \sum_{i=1}^{n} \delta(\mathrm{POS}(w_i), p_1), \dots, \sum_{i=1}^{n} \delta(\mathrm{POS}(w_i), p_{49}) \right]^T \in \mathbb{N}^{49}$$

In the above, $\mathrm{POS}(w_i)$ is the spaCy 3.5.0 function that returns the pos-tag of the token ($w_i$). Thus, the vector $\mathbf{u}(s) = [u_1(s), u_2(s), \dots, u_{49}(s)]^T$ is a 1×49 matrix of its individual counts, where each $u_j(s)$ is how many tokens in *s* were tagged as the jth POS tag. δ is the Kronecker delta function. $\delta = 1$ when $\mathrm{POS}(w_i) = p_j$ and $\delta = 0$ otherwise.

**6.2 Stage Three: Dependency Inventory Vector**

A further innovation of our method vectorizes dependency tags into an inventory vector. The dominant temporal meaning of a sentence derives from multiple intra-constitutive relationships. A core principle of dependency grammar is that the meaning of a sentence resides within non-nested relational adherents (Halliday 1994). This basic principle influenced our use of dependency labels in order to provide a mechanism for a classifier to analyze these intraconnected relationship. spaCy 3.5.0 offers 45 dependency labels that can be used for dependency parsing. In alignment with our vectorization of part-of-speech tags, we obtained a vector representation for each of the 45 spaCy 3.5.0 dependency labels (Honnibal et al. 2020). This gave a 45-dimensional vector space, with one dimension for each dependency label. At each element in the Dependency Inventory Vector, we set the value to the number of times that a corresponding dependency tag occurs in a sentence. Importantly, this means that if a dependency label is not found in a sentence, it is counted as a zero in its corresponding element of the feature vector. Each vector, therefore, specifies the dependency labels that describe the sentence as well as those that do not. In this way, every element of the vector encodes information about the sentence.

The above process is represented by the following equation that represents the dependencies by themselves in a 45-dimensional vector space (see below 6.4 for the equation representing the combination of the dep and pos image vectors together). As above, ***s*** consists of an ordered list of tokens, such that $w_1$ is the first token and $w_n$ is the last: $\mathbf{s} = [w_1, w_2, \dots, w_n]$. Let d(*s*) = the fixed, ordered inventory of all of the 45 spaCy 3.5.0 defined fine-grained dependency tags: $d(s) = [d_1(s), \dots, d_{45}(s)]$. Using the 45 spaCy functions, we then get:

$$\mathbf{d}(s) = \left[ \sum_{i=1}^{n} \delta(\mathrm{DEP}(w_i), q_1), \dots, \sum_{i=1}^{n} \delta(\mathrm{DEP}(w_i), q_{45}) \right]^T \in \mathbb{N}^{45}$$

In the above, d(s) is the dependency-count vector for sentence s. It has 45 elements, one for each spaCy 3.5.0 dependency label. The j-th component, $d_j(s)$, specifies how many times the j-th dependency label appears in s. DEP($w_i$) is spaCy's dependency label for token $w_i$. $q_j$ is the j-th tag in our fixed list of 45 dependency tags. δ(a, b) is the Kronecker delta, where a=DEP($w_i$) and b=$q_j$. δ(DEP($w_i$),$q_j$) = 1 when token ($w_i$)'s dependency label equals $q_j$ , 0 if otherwise.

### 6.3 Method Development Stage Four: Appending Futurate Inventory Vectors

The futurate can be especially difficult for programmatic methods. To address this problem, we inventory sentences in our dataset for seven chosen futurate forms. We developed 7 Python scripts to capture the corresponding futurate forms. Each script generates a corresponding feature in the Futurate Inventory vector space. At each element in a vector, we set the count value to the number of times that each script function detected a futurate form. If a specific futurate form was not detected, then a zero is inserted at the corresponding element of the vector. In this way, every element of the vector encodes futurate information about the sentence. (See Appendixfor examples of Specific Futurate forms.)

The above process can be represented in the following equation. As above, let s = [$w_1$, $w_2$, … , $w_n$], where n denote the number of tokens in the sentence **s**. Each token $w_i$ is annotated with part-of-speech tags. We define a set of N = 7 functions that correspond to seven distinct futurate constructions. Using the 7 different futurate functions, we get the following:

$$f(s) = \left[ \sum_{i=1}^{n} \delta(\mathrm{FUT}(w_i), r_1), \; \ldots, \; \sum_{i=1}^{n} \delta(\mathrm{FUT}(w_i), r_7) \right] \in \mathbb{N}^7$$

In the above, f(s) is the futurate feature vector for the whole sentence. $\mathbb{N}^7$ denotes the 7-dimensional space of natural number vectors.

### 6.4 Stage Five: Combining POS, Dependency and Futurate Vectors

Combining the 49 fine-grained pos tags (u(s)) with 45 dependency labels (d(s)) yields a 94-dimensional vector space. To this combination, we add 7 more dimensions using the 7 futurate function (f(s)). In total this yields a unified 101-dimenional morphosyntactic vector space, wherein each sentence has a representation that we term a "*Morphosyntacton*". This is represented by the following equations:

$$u(s) = [u_1(s), u_2(s), \ldots , u_{49}(s)]$$

$$d(s) = [d_1(s), \ldots , d_{45}(s)]$$

$$f(s) = [f_1(s), \ldots , f_7(s)]$$

$$u_j(s) = \sum_{i=1}^{n} \delta(\mathrm{POS}(w_i), p_j),\ j = 1, \ldots, 49$$
$$d_j(s) = \sum_{i=1}^{n} \delta(\mathrm{DEP}(w_i), q_j),\ j = 1, \ldots, 45$$
$$f_j(s) = \sum_{i=1}^{n} \delta(\mathrm{FUT}(w_i), r_j),\ j = 1, \ldots, 7$$

Combining the u(s), d(s), and f(s) inventory vectors of a sentence, we get msi(s), a global morphosyntactic inventory vector of the sentence s:

$$\mathrm{msi(s)} = \begin{bmatrix} \mathrm{u(s)} \\ \mathrm{d(s)} \\ \mathrm{f(s)} \end{bmatrix}^{T} \in \mathrm{N}^{101}$$

**6.5 Stage Six: *XGBoost Classifier***
Our choice of XGBoost primarily derived from its lack of a scaling requirement (Chen and Guestrin 2016). Since our vectorized feature space contains counts of individual features, there is the intuitive concern that scaling our vectors within fixed intervals would deprive a classifier of granularity. 1799 sentences from our annotated dataset were put into a pandas dataframe and used to train the classifier.

## III. RESULTS AND DISCUSSION

**7.0 Accuracy**
75% of our pandas dataframe was used for training on XGBoost and 25% for testing. No hyperparameter tuning was done; default values were used. Below, see our classification report, confusion matrix, multi-log loss chart, classification error chart SHAP values.

**Table 6: Classification Report**

| Class | Precision | Recall | F1-score | Support |
|---|---|---|---|---|
| Past | 0.95 | 0.94 | 0.94 | 155 |
| Present | 0.91 | 0.89 | 0.90 | 142 |
| Future | 0.89 | 0.93 | 0.91 | 153 |
| Accuracy | | | 0.92 | 450 |
| Macro avg | 0.92 | 0.92 | 0.92 | 450 |
| Weighted avg | 0.92 | 0.92 | 0.92 | 450 |

The classifier achieved an overall F1-score of 0.92. Outside the very strong performance in the past category, performance for the present and future categories approached past category accuracy levels. As mentioned above, (see **Introduction**) this is particularly noteworthy, since prior studies have consistently reported substantially lower performance for the present and future categories.

**Figure 4: Confusion Matrix for XGBoost**

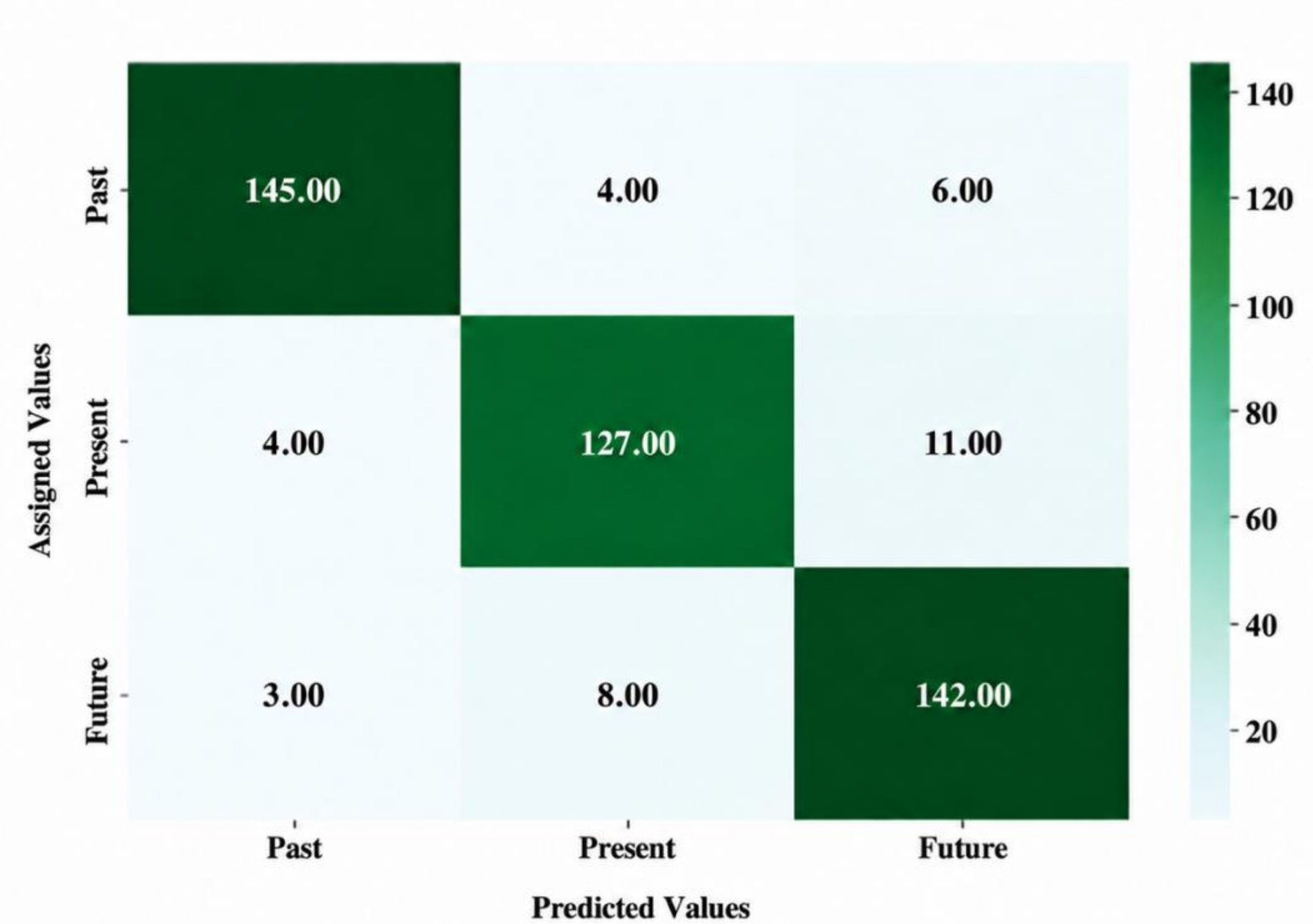

The confusion matrix indicated strong classification performance across all categories, with only minor misclassifications. Of the 155 sentences labeled as past, 145 were correctly identified, while 4 were misclassified as present and 6 as future. For the present category, 127 sentences were correctly classified, with 4 misclassified as past and 11 as future. For the future category, 142 sentences were correctly classified, with 8 misclassified as present and 3 as past. Overall, the majority of errors occurred between the present and future categories, with the present being the most problematic.

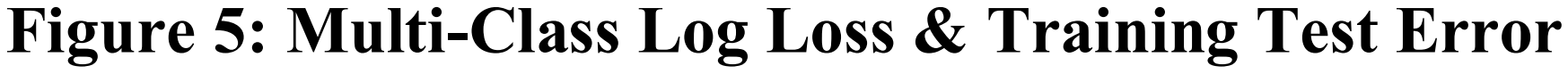

**Figure 5: Multi-Class Log Loss & Training Test Error**

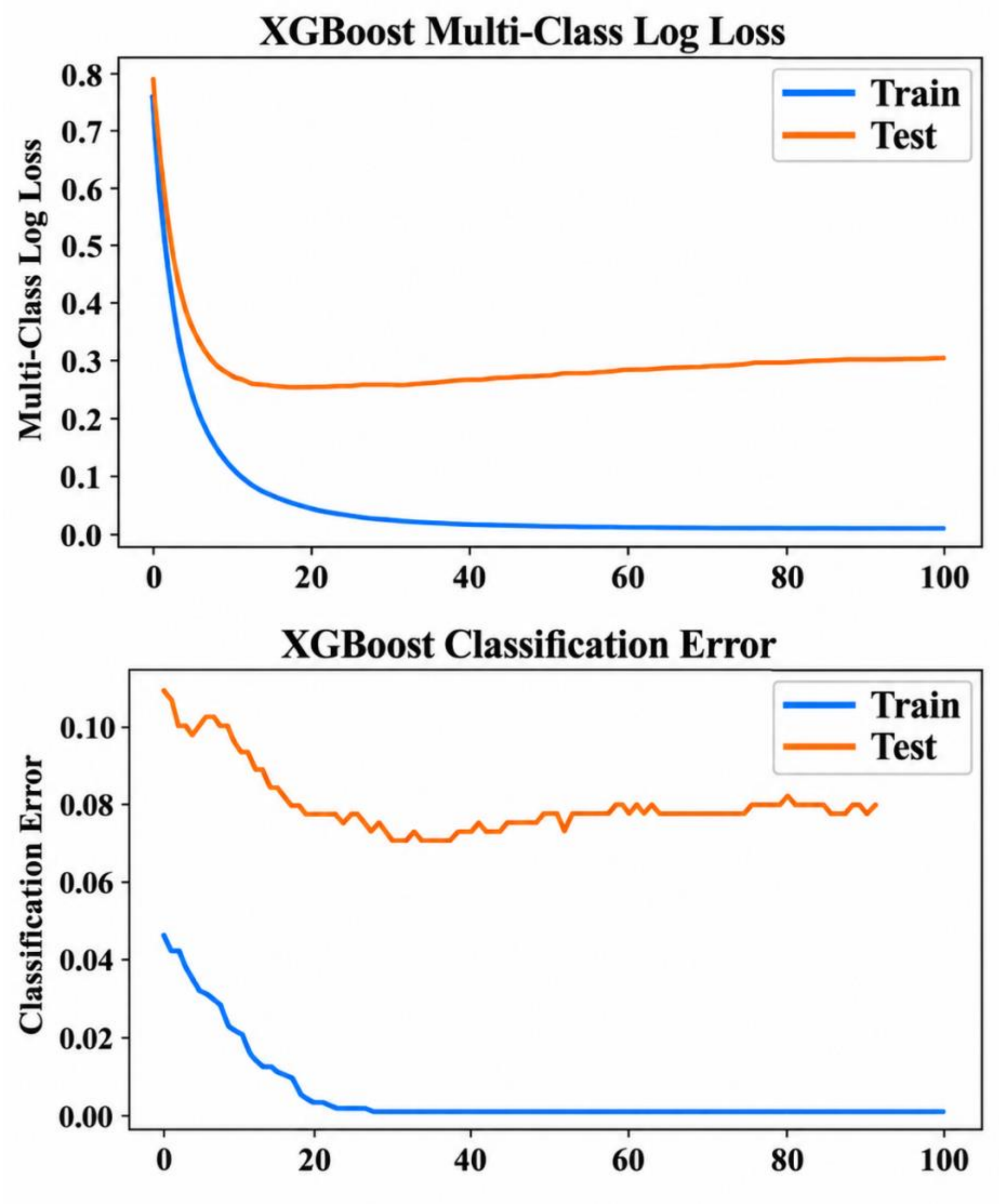

The multi-class log loss decreased consistently for both training and testing sets, indicating effective learning and stable generalization. Although early stopping was not applied, the steady downward trend of both curves suggests reliable convergence. At the final iteration, the training log loss approached ~ 0.05, while the test log loss stabilized around ~ 0.3. The classification error curves provide further insight. Training error decreased to nearly 0.00, while the test error stabilized around 0.08. The resulting gap between training and test errors reflects mild overfitting; however, cross-validation scores, with a low standard deviation (listed below), suggest that this effect is not substantial. Keeping in mind that a degree of overfitting is highly expected in this problem domain, as temporal text classification often involves overlapping linguistic cues between categories. Nevertheless, the model demonstrates strong classification performance with the suggestion of a limited generalization error.

**Table 7: Cross-Validation Scores for XGboost**

| **Scores** | 0.9250, 0.9083, 0.9166, 0.9330, 0.8711 |
|---|---|
| **Mean** | 0.9110 |
| **C-V Standard Deviation** | 0.02128 |

### 7.1 Top-Ranked SHAP Features

Table 8 presents the top-ranked features based on mean absolute SHAP values (Lundberg and Lee 2017; Shapley 1953). The SHAP value of a feature indicates its contribution to the output (prediction) of a classifier. The highest-ranked features are dominated by verb tense markers (e.g. VBD, VB, VBP, VBN), as intuitively expected. These are followed by modal features (MD) and syntactic dependencies. Futurate markers (e.g. fu_3) also appear among the top features. Less intuitive, is the influence of features that do not obviously represent time related concepts, e.g. NNP (0.162308, Proper Noun Singular) or PRP (0.156923, Personal Pronoun). This suggests the possibility of hidden or at least non-obvious relationships between morphosyntactic features and the temporal dimensions of sentences (See Appendix A.20 for complete table).

**Table 8: Numerical SHAP Values**

| Rank | Feature | Importance |
|---|---|---|
| 1 | VBD | 1.664185 |
| 2 | VB | 0.748659 |
| 3 | MD | 0.503552 |
| 4 | VBP | 0.379765 |
| 5 | VBN | 0.378118 |
| 6 | punct | 0.255277 |
| 7 | amod | 0.211524 |
| 8 | VBZ | 0.200538 |
| 9 | fu_3 | 0.1822 |
| 10 | NNP | 0.162308 |
| 11 | PRP$ | 0.159578 |
| 12 | PRP | 0.156923 |
| 13 | aux | 0.155046 |

| 14 | TO | 0.153113 |
|---|---|---|
| 15 | advcl | 0.123179 |
| 16 | advmod | 0.11803 |
| 17 | JJ | 0.099034 |
| 18 | NN | 0.097591 |
| 19 | ccomp | 0.093973 |
| 20 | compound | 0.093768 |

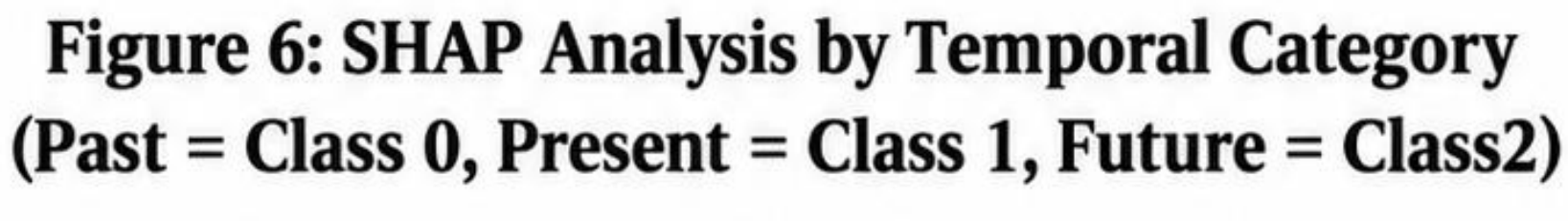
Figure 6: SHAP Analysis by Temporal Category
(Past = Class 0, Present = Class 1, Future = Class2)
VBD
VB
MD
VBN
VBP
punct
VBZ
amod
aux
PRP
NNP
fu_5
JJ
TO
nsubj
compound
NN
poss
NNS
pobj
Class 0
Class 1
Class 2
0
1
2
3
4
mean(|SHAP value|) (average impact on model output magnitude)

For the past class, VBD (past tense verb) was the dominant predictor, with VBN (past participle) also providing major support. VB emerged as the most important feature for the present whereas VBD and MD (Modals) also make major contributions. However, except for VBN, all other elements make some contribution. Future classification, in contrast, was most strongly associated with VB and MD, as expected for modal constructions (e.g., *will + VB, shall + VB*). Surprisingly, NNP (proper nouns) and NNS (plural nouns) PRP (personal pronouns) contribute to the present classification even though by themselves they appear to be neutral with respect to time. This suggests that morphosyntactic features can contribute to a temporal category even though they are not obviously related to time. While the contribution of a single such feature may be modest, the total contribution of multiple such features can be significant (see Feature Ablation Curve below).

**7.2 Analysis of SHAP Feature Contributions to Classifier Accuracy**

Although SHAP analysis can be used to rank features by their influence on a model, it cannot tell us by itself how accuracy is directly affected by a feature, particularly whether it is necessary or whether removing it hurts accuracy. Ablation techniques give us the possibility of determining directly how accuracy changes in regards to removing individual features or limiting the number of features. Ablation analysis was performed using the SHAP-derived feature ranking. Specifically, features were ordered by their mean absolute SHAP values, and subsets of the top N features were incrementally used to retrain and evaluate the model. Accuracy was then measured as a function of the number of included features. Results show that model accuracy increases by including the top-ranked features up to approximately 40 features, after which performance plateaus. This indicates that mid-ranked features contribute meaningfully to performance, while the long tail of lower-ranked features provides negligible additional benefit, at least with the data set used in this study.

**Figure 7**

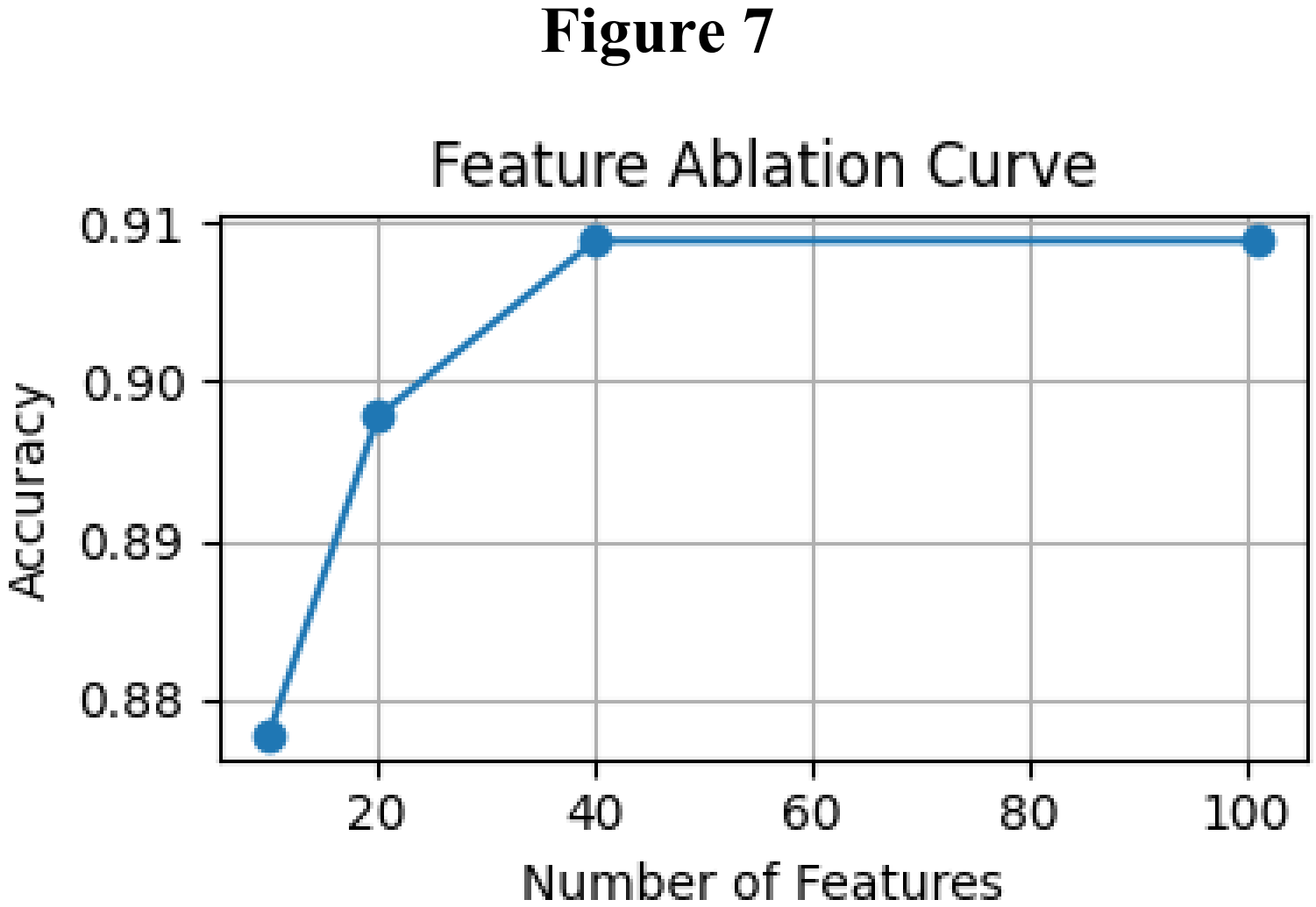

**7.3 Results from Adding Pretrained Sentence Embeddings from Google's Universal Sentence Encoder (GUSE) and Sentence-BERT**

What is the influence of adding pretrained embeddings to our dataframe? To address this question, we generated pretrained embeddings for the same set of 1,799 sentences used to produce the XGBoost results discussed above. This process was carried out in two stages using two different text embedding models: GUSE (512 dimensions, pretrained) and Sentence-BERT (768-dimensions, pretrained**).** We evaluated the impact of (**1**) concatenating each set of pretrained embeddings individually with our primary dataframe (morphosyntactic features) and (**2**) measuring the accuracy of each set of embeddings by themselves on our dataset. XGBoost was the primary classifier in all cases. Default hyper parameters without tuning were used. In the figure below, our primary model is designated as ***POS-dep + Futurate (vectors)***. The inclusion of the 512-GUSE embedding is designated as ***POS-dep + Futurate + GUSE***. The inclusion of 768-BERT embedding is designated as ***POS-dep + Futurate + BERT***. BERT by itself on our dataset is designated as ***BERT only***. GUSE by itself on our dataset is designated as ***GUSE only***.

## Figure 8: Accuracy and Cross-Validation Performance

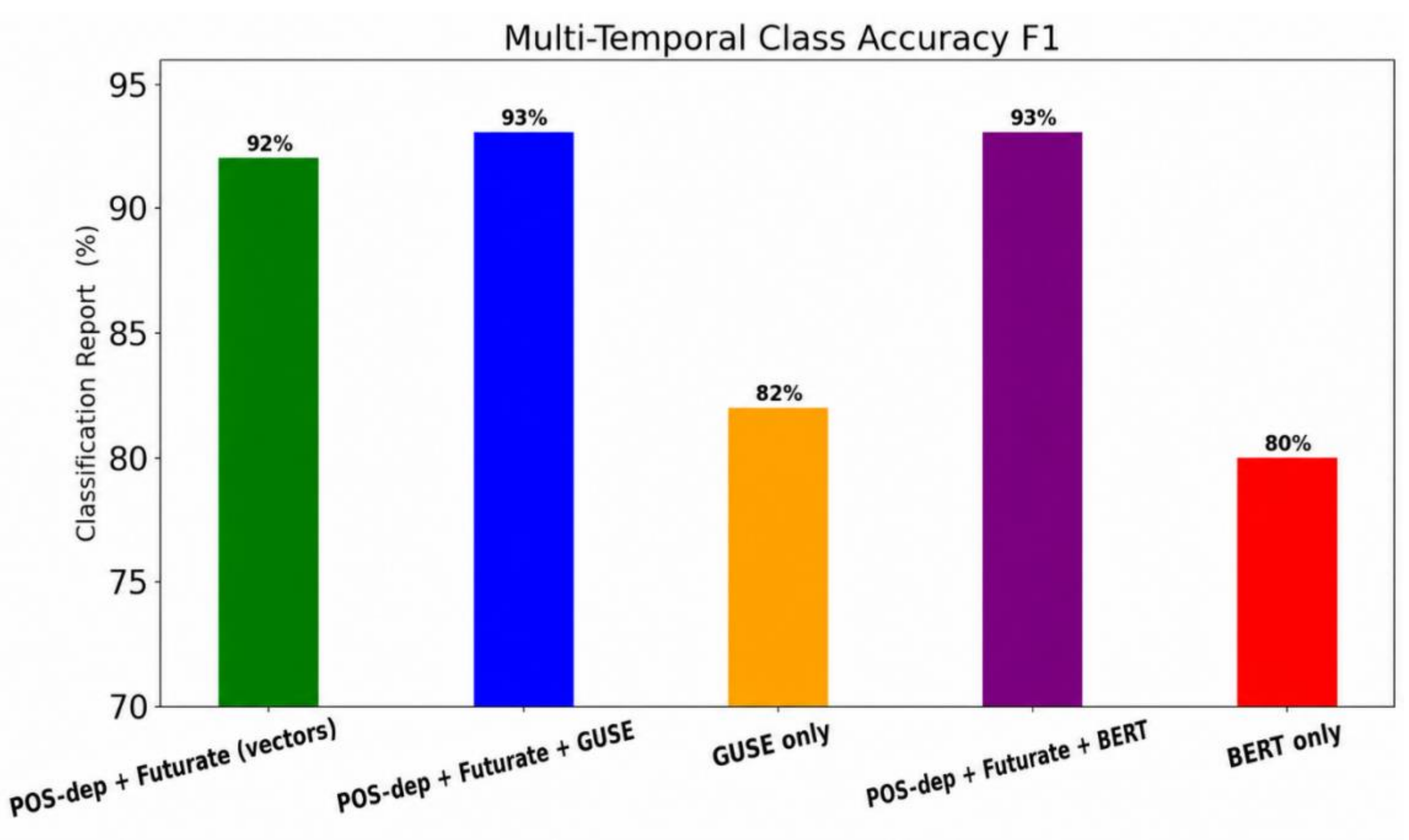

**Note**: Here, *pp* denotes the absolute change in accuracy measured in percentage points.

The mean cross-validation score for our baseline classifier using POS and dependency features (POS-dep + futurate) achieved a multiclass accuracy of 91.1% on the held-out test set. Incorporating GUSE embeddings increased performance to 93.1%, representing a 2.0 absolute point change. By contrast, adding BERT embeddings raised accuracy only slightly to 91.7%, an improvement of just 0.6 %. Although the cross-validation standard is tightly clustered around the mean for (POS-dep + futurate) in absolute terms, the inclusion of GUSE to the dataframe generates an approximate 1/3 decrease in the standard deviation. The use of embeddings by themselves - GUSE only or BERT only - shows that our model outperforms pretrained embeddings by approximately 12 percent for BERT-Only and 10 percent with GUSE-Only.

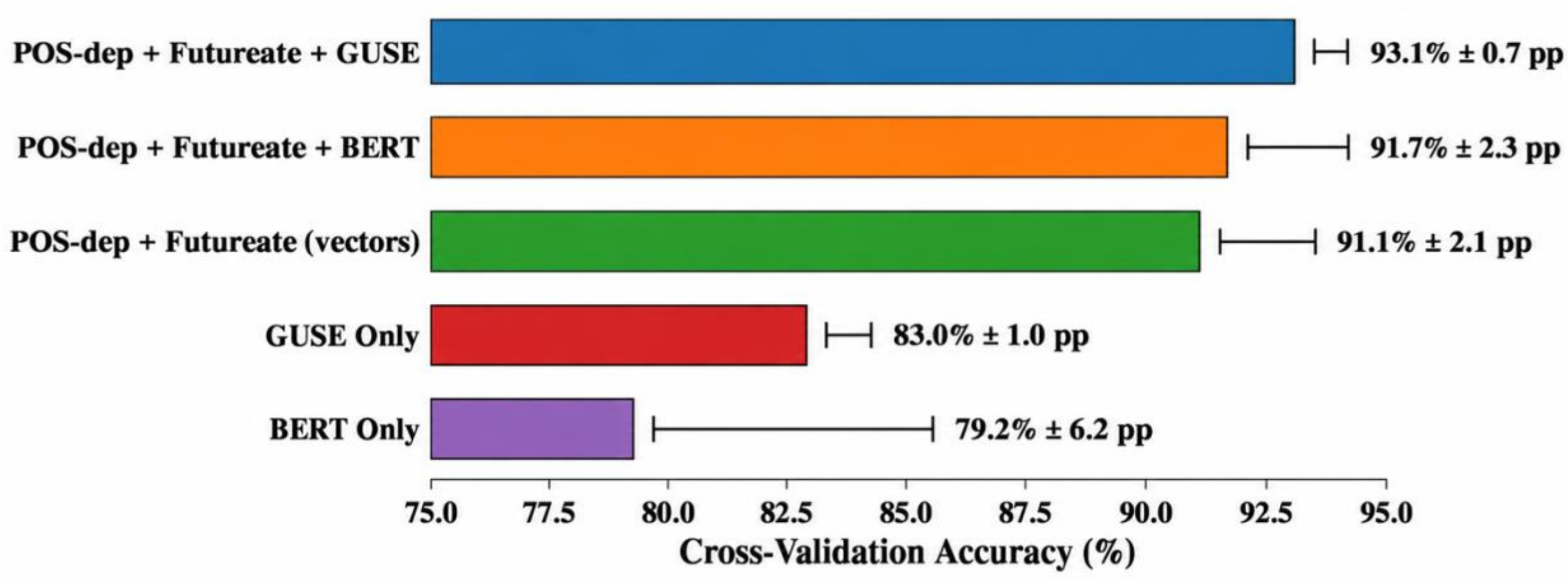

### 7.4 Accuracy by Temporal Category

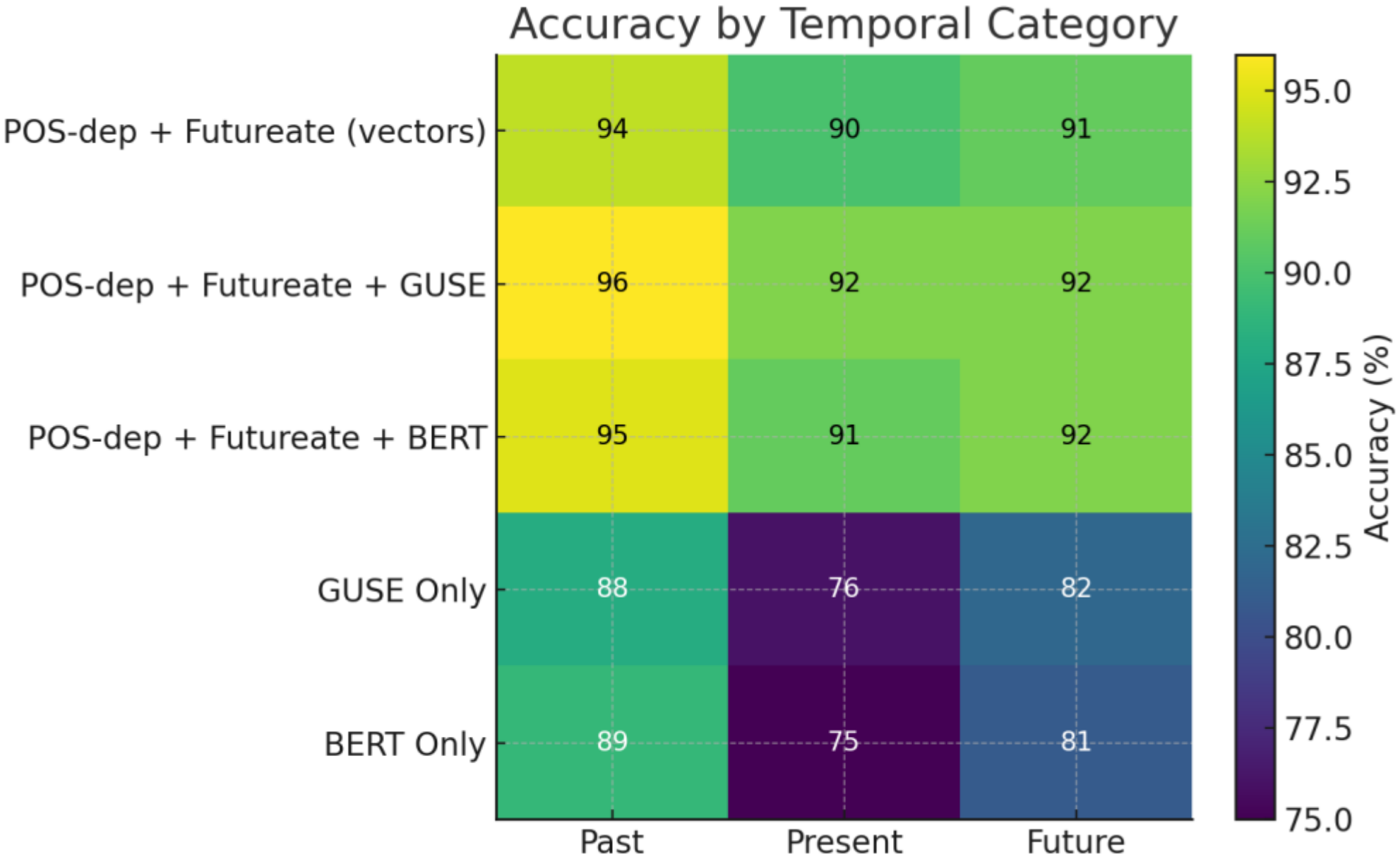

The ***POS-dep + Futurate + GUSE*** configuration produced the largest positive deltas across all temporal categories—approximately +2.5 % for Past, +2 % for Present, and +1 % for Future. ***The POS-dep + Futurate + BERT*** system achieved smaller but consistent gains (+1 – 1.5 %). In contrast, *GUSE* and *BERT* individually performed substantially worse, falling as much as 15 % below baseline accuracy, with the most pronounced declines observed in the Present or Future category. These results confirm that augmenting explicit syntactic-temporal features with contextual embeddings modestly enhances performance, while both *GUSE* and *BERT* embeddings in isolation remain comparatively inadequate for temporal classification.

### 7.5 Results: Confusion Matrix by Temporal Category Comparison

Below are two tables that separately compile the correct vs misclassifications from five different confusion matrices.

**Figure 10: Comparison of Confusion Matrix Correct Classifications by Temporal Category**

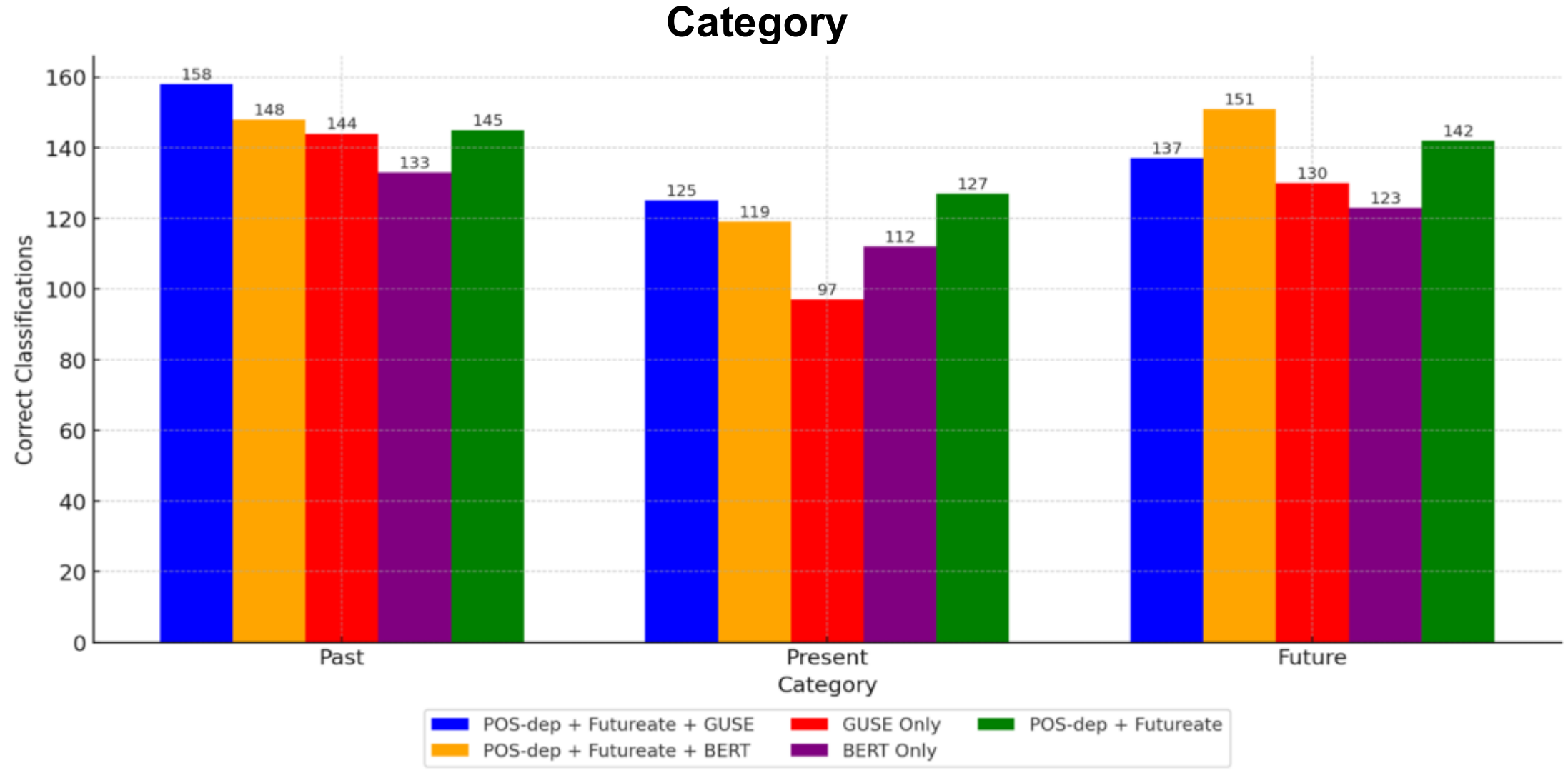

**Figure 11: Comparison of Confusion Matrix Misclassifications by Temporal Category**

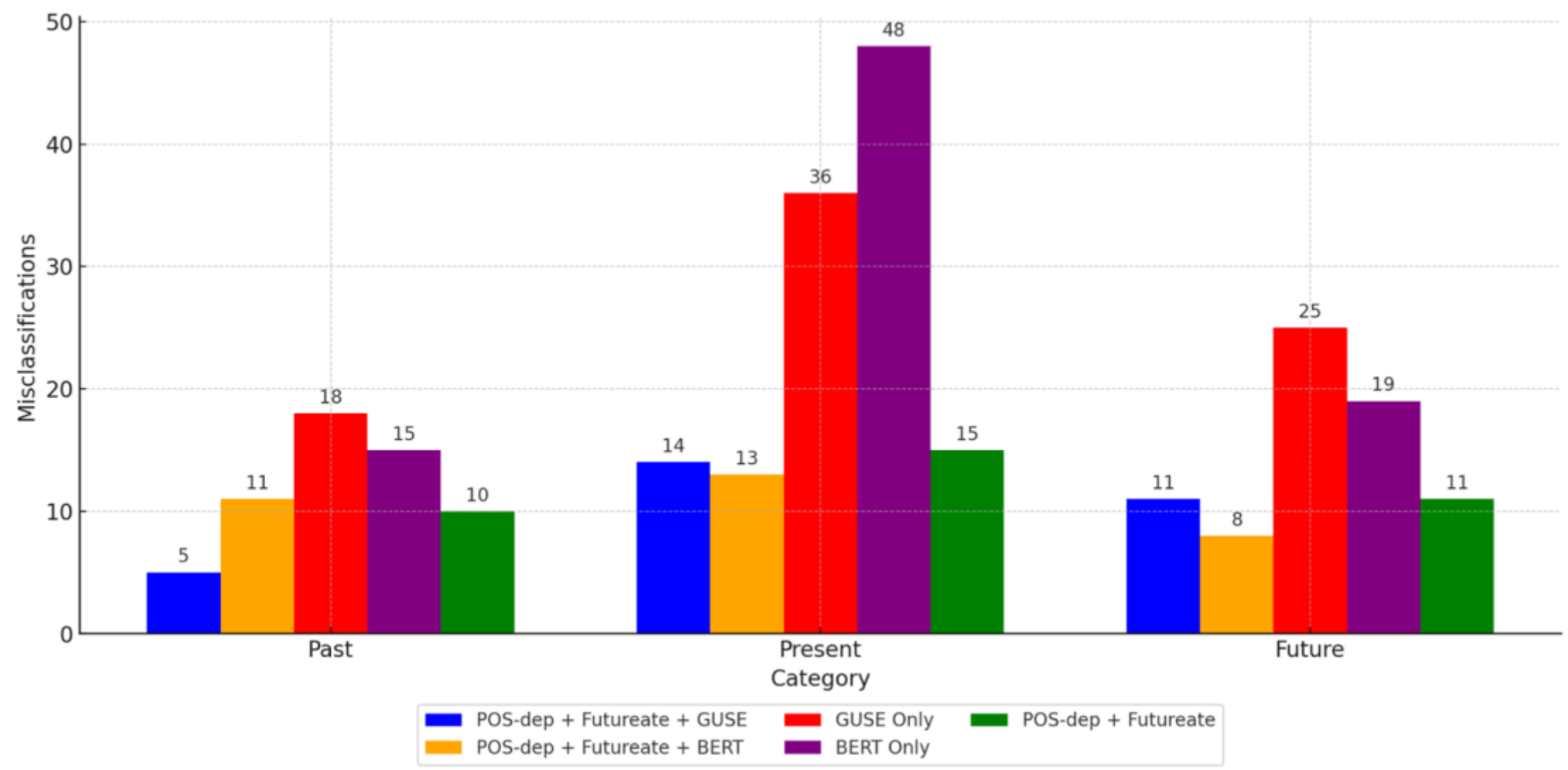

Our Baseline model, **The POS-DEP + FUTURATE**, achieves high performance across all three temporal categories with small, class dependent differences relative to the alternative models in terms of correct classifications and misclassifications. For the past category (*Class 0)* and the future (*Class 2*) the inclusion of the **GUSE** embeddings (**POS-DEP + FUTURATE + GUSE)** generates more correct classifications and fewer misclassifications than the baseline, approximately 1.21% in absolute percentage points. However, for the present class (*Class 1*) our baseline model produces the most correct classifications and the fewest misclassifications. Therefore, our baseline model outperforms the **POS-DEP + FUTURATE + GUSE** model by ≈1.20 percentage points on the present category. The **POS-DEP + FUTURATE + BERT** model performs slightly worse than our baseline model across all classes. **GUSE only and BERT only** show significantly fewer correct classifications and higher misclassification rates, approximately 6.55-11.72 percentage points lower. Therefore, gains from the inclusion of the GUSE embeddings to our baseline model are modest. This suggests that hybrid gains derive primarily from the syntactic precision of the **POS-DEP + FUTURATE** framework.

**7.6 Results: Loss Trends and Overfitting Comparison Across Models**

The figure below compares **model learning and generalization behavior** for the baseline (**POS-dep + futurate**) and the two hybrid models (**POS-dep + futurate + GUSE** and **POS-dep + futurate + BERT**), as well as the two pure pretrained embedding modules: **GUSE only**, **BERT only**. Below, you see the evolution of training, validation, test, and multi-log losses across epochs.

**Figure 12**

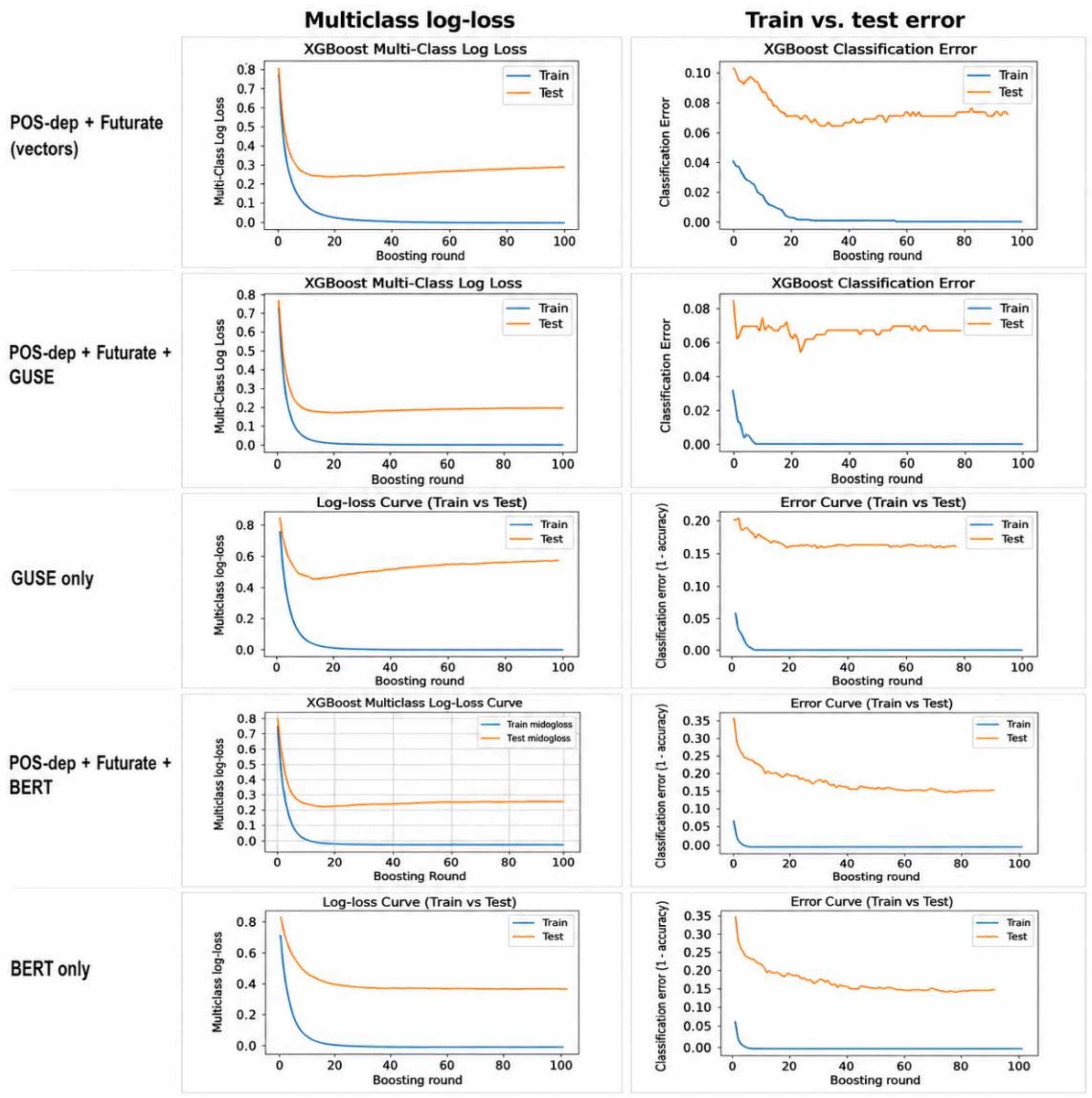

In the above figure, considering both multi-class log-loss and train versus testing: **POS-DEP + FUTURATE + GUSE** shows the least overfitting, followed by POS-dep + Futurate baseline. POS-dep + Futurate + BERT shows moderate overfitting. The embedding only models perform worse overall. Of these, **GUSE only** shows the strongest overfitting pattern, because its test loss and test error diverge more clearly from the training curves. **BERT only** also performs weakly overall, but its curve suggests poorer generalization more than pronounced overfitting.

**8.0 Study's Limitations**

First, our analysis is restricted to sentence-level temporal reference and does not account for temporal interpretation across larger discourse units. In natural language use, temporal reference is often shaped by prior context that occurs as a result of accruing narrative progression, and discourse-level anchoring.

Second, our annotation of temporal reference into discrete categories—past, present, and future— simplifies the granular complexity of temporal references. For example, our annotation policy does not explicitly differentiate conditional references or differentiate between the near and distant future.

Third, our inventory of Sentence-BERT **(**768-dimensions**)**, in comparison with our POS-dep + Futurate Vectors model (101-dimensions), should be interpreted with some care. No task specific fine-tuning of Sentence-BERT's internal transformer parameters was performed.

## IV. CONCLUSION AND FUTURE WORK

This study makes two primary innovations. The first innovation is the introduction of the Morphosyntacton, a unified vector representation used for sentence-level temporal classification that allows sentences of varying lengths to be represented without padding. This representation suggests that the global temporal reference frame of syntactically complex sentences can be modeled through explicit structural features without the presence of pretrained embeddings. These features include part-of-speech information, dependency information, and futurate pattern counts. The accuracy results show that this structural framework achieves an overall multiclass accuracy of 92 % on syntactically complex sentences. Performance was strongest for past classification, but the model also remained comparatively strong in the present and future categories.

The other point is that concatenating pretrained embeddings with our primary feature space produces only modest average gains in accuracy. Used alone, however, GUSE and BERT underperform the baseline by roughly 10 and 12 absolute percentage points, respectively. The clearest hybrid-positive effect appears in the GUSE hybrid through reduced overfitting and tightening the standard deviation of the cross-validation.

A further important point is our finding that morphosyntactic features can contribute to a temporal category even though they are not obviously related to time. This finding suggests the possibility of hidden or at least non-obvious relationships between morphosyntactic features and the temporal semantics of sentences.

A second innovation lies in the annotation framework. The study introduces a more explicit policy for assigning dominant temporal intent in temporally complex sentences. This policy relies on exemplars and a clearer definition of dominant temporal orientation. It could therefore be argued that the annotation procedure forms part of the study's contribution itself, rather than serving merely as a preliminary stage. In addition, we describe a programmatic method for the generation of automatically annotated training samples that correspond to different temporal categories.

Future work may benefit by focusing on the level of generalizability present in our proposed model beyond our current data set and task conditions. Specific emphasis could focus on the generalizability across different genres, registers and domains of varying types of text sources. The other future point is determining whether our model can be extended beyond the sentence level. Specifically, applied to discourse level modeling of temporal references that can occur through accruing narrative progression in textual sources. Finally, another possible agenda would be the application of our method to segmented clauses, when dependency parsers become sufficiently accurate. This possibility might offer the capacity to characterize not only the net counts of temporal references in a large document but also the sequences of temporal references, adding another dimension to the analysis of temporal references (Cleveland, 2023: 234-236).

# Appendix

**A.1 Measuring Shannon Entropy**

$$p(w) = \frac{f(w)}{\sum_{v \in V(C)} f(v)}$$

$$H(C) = -\sum_{w \in V(C)} p(w) \log_2 p(w)$$

In the above, C = complete corpus, V(C) = set of unique word types in the corpus
w = a particular word type, v = summation index over word types, f(w) = frequency of word w,
p(w) = probability of word w, H(C) = Shannon entropy of the corpus
log2 = base-2 logarithm, bits = unit of entropy.

The corpus was processed in Python using SpaCy 3.5.0. Each sentence was tokenized into individual word tokens. Python was then used to count how many times each distinct word type occurred across the complete corpus. The frequency of each word was divided by the total number of word tokens to obtain its probability. These probabilities were then entered into the Shannon entropy equation.

**A.2 Average Words per Sentence Across the Corpus**

This measure represents the average number of words per sentence across the entire corpus. **L(s)** = number of words in a single sentence (s). **N** = total number of sentences in the corpus. The summation adds the length of every sentence. Dividing by N gives the average number of words per sentence across the corpus.

**A.3 The Average Number of Characters per Sentence Across the Corpus**

$$\text{AvgChars} = \frac{1}{N}\sum \chi(s)$$

N is the total number of sentences in the corpus. **s** denotes single sentence**. χ(s) is** the character count of sentence *s*. This includes all letters, spaces, and punctuation exactly as they appear in the sentence. This provides a measure of sentence length independent of tokenization, capturing structural size purely in terms of characters.

**A.4 Average Clauses per Sentence**

Average number of clauses per sentence across the corpus. N is the total number of sentences in the corpus. **s** denotes a single sentence. ClauseCounts(**s**) refers to the number of clauses identified in a sentence using SpaCy 3.5.0 dependency labels. This quantifies how many clause structures typically appear within sentences, reflecting the degree of syntactic embedding and complexity in the corpus:

$$\text{AvgClauses} = \frac{1}{N}\sum \text{ClauseCount}(s)$$

**A.5 Average Number of Dependency Relations per Sentence.**

Average number of dependency relations per sentence across the corpus. For each sentence, s, spaCy 3.5.0 identifies a set of dependency relations. Each dependency relation, or dependency arc, represents a syntactic link between two words. D(s) the number of dependency labels or parsed dependency relations in sentence s. The corpus-level average is calculated by adding D(s) for all sentences and dividing the total by N, the total number of sentences in the corpus:

$$\text{AvgDeps} = \frac{1}{N}\sum_{s=1}^{N} D(s)$$

$$D(s) = |\text{Dependencies}(s)|$$

**A.6 Subordination-to-Coordination Ratio**

Both subordinate and coordinated clauses were identified using spaCy 3.5.0 dependency labels, with coordination specifically determined by tokens labeled conj. Here, s denotes a single sentence, and N is the total number of sentences in the corpus. S(s) is the number of subordinate-clause relations identified in sentence s, and C(s) is the number of coordination relations identified in sentence s:

$$\text{SubCoordRatio} = \frac{\sum_{s=1}^{N} S(s)}{\sum_{s=1}^{N} C(s)}$$

**A.7 AvgPreps**

**AvgPreps =** The average number of prepositional relations per sentence across the corpus. N is the total number of sentences in the corpus. (s) is a single sentence. P(s) is the number of tokens in a sentence that is determined by a corresponding dependency label = prep:

$$\text{AvgPreps} = \frac{1}{N} \sum_{s=1}^{N} P(s)$$

**A.8 Avg Passives**

The average number of passives. The sentence is parsed using spaCy's dependency parser to inventory passives. A counter is incremented whenever a token has a dependency label of: auxpass (passive auxiliary, e.g., was written) or nsubjpass (passive nominal subject, e.g., The book was written). N = total number of sentences. P(s) = passive markers in a sentence:

$$\text{AvgPassives} = \frac{1}{N} \sum_{s=1}^{N} P(s)$$

**A.9 Avg Tree Depth**

Measures the average tree depth**. $D_{depth}$(s)** is defined as the maximum number of hierarchical dependency levels from the ROOT to any descendant token in sentence (s). The maximum depth of the Add the tree depth value for every sentence dependency tree for sentence (*s*). This is computed by: Identifying the ROOT node of the sentence. Next, measuring the longest path from the ROOT to any descendant token. This is followed by Traversing the dependency tree, which measures the longest path from the ROOT to any descendant token. ∑ s=1 N means add the tree depth value for every sentence. N is the total number of sentences.

$$\text{AvgTreeDepth} = \frac{1}{N} \sum_{s=1}^{N} D_{\text{depth}}(s)$$

## A.10 Average Subjects per Sentence

$$\text{AvgSubjects} = \frac{1}{N} \sum_{s=1}^{N} \text{SubjectCount}(s)$$

SubjectCount(s) is the number of tokens in sentence s that spaCy labels as nsubj or nsubjpass. The nsubj label identifies an active nominal subject, while nsubjpass identifies a passive nominal subject. After dependency parsing, spaCy stores each token's readable dependency label in token.dep_, and each matching token is counted once. N means the total number of sentences in the corpus. The summation symbol means that the subject counts are added across every sentence. The combined subject count is divided by N to produce the average number of subjects per sentence.

**A.11 Annotation Policy for Futurate and Future Idioms**

The other type of exemplars used for future classified sentences involved the futurate and future idioms. Futurate sentences, where the present tense is used to reference the future, can be challenging for raters. This challenge can occur with or without temporal adverbs. Some examples that were used to help navigate this complexity included sentences without modals and future idioms:

**Table A1 Exemplars of Futurate and Future Idioms Sentences for Future**

| **Sentence** | **Classification** |
|---|---|
| He is going to buy a car. | Future |
| We are buying a car tomorrow. | Future |
| There is a long road ahead. | Future |

**Table A2 Exemplars of Futurate and Future Idioms Sentences for Future**

| **Futurate Types** | **Sentence/Word Examples** |
|---|---|
| Present continuous: the use of a present tense verb coupled with a present participle | I am going to buy a car.<br>I am not going to buy a car. |

| The use of an auxiliary helper verb coupled with an action verb | I will buy a car.<br>I will not buy car |
| --- | --- |
| The use of the verb 'have' followed by a preposition 'to' and an infinitive verb: | I have to buy a car. |
| The use of an auxiliary modal aux verb: 'will' coupled with the verb 'to be' | I will be married. |
| The use of a modal verb followed by a base verb: | She will sing beautifully |
| The use of the following two auxiliary verbs: 'have' or 'ought' followed by a preposition and a variable verb. | I have to buy a car.<br>I ought to drive a car. |
| Sentences containing the future perfect | I will have completed my degree |

**A.12 Annotation Policy for Past References**

Unlike the present and future, syntactical references to the past can offer semantic clarity and accessibility. This is likely one of the reasons that past category machine learning classification accuracy scores are usually much higher in comparison to the present and future. One reason for this is that most verbs in English usually have a tight morphological linkage with some sort of inflectional change. Tight morphological linkages with suffix changes and helper verbs likely make it easier for a classifier to classify corresponding past references. One major exception to this linkage occurs in past phrases with adverbial functions. Below are some of the examples used to guide past annotations and sentences generated from our past-template.

**Table A3 Exemplars Sentences Used for Past Designation.**

| Sentences | Tense | Model Classification |
| --- | --- | --- |
| I had some time before I left work. | Simple Past | Past |
| I took some time to find parking | Simple Past | Past |
| By the time I arrived at the party, everyone had already left. | Past Perfect | Past |
| In a previous era. | Past Phrase | Past |

| Sentences | Tense | Model Classification |
|---|---|---|
| I have spent a lot of time planning my approach. | Present Perfect | Past |
| We've increased spending for discretionary programs by a very responsible 4 percent, above the rate of inflation. | Present Perfect | Past |
| My fellow Americans, we have crossed the bridge we built to the 21st century. | Present Perfect | Past |
| For two years in a row, Congress has supported my plan to hire 100,000 new qualified teachers to lower class size in the early grades. | Present Perfect | Past |

**A.13 Annotation Policy for Future References**

The benchmark examples used for our future annotation policy reflect what linguists term 'futurity" (Huddleston et al. 2021:75). This process can be highly tied to modal auxiliaries and semi-modals. Modal auxiliaries can also be separated between the *epistemic* and the *deontic*, where deontic modals express notions of necessity, e.g. "The storm must be over soon." (Huddleston et al. 2021:70). Epistemic modals tend to express what is possible, e.g. "They should try again (Huddleston et al. 2021:70)." More broadly, modal auxiliaries are known to especially reference the future through what is called "irrealis" (Palmer 2001:185–202). By this, we mean it expresses "possibility" (Frawley 2008:203), or more broadly "futurity" (Palmer 2014:137). More concretely, futurity can refer to "matters beyond the factual here and now" (Papafragou, Trueswell, and Gleitman 2022:152). Our annotation policy held to the broad tenet that sentences which reflect a type of intentionality that concerns what is not in the past or not present (i.e. a future desire or necessity), should be classified as being a future oriented sentence. This policy held for sentences irrespective of whether modals were present. Below are some examples:

**Table A4: Model Sentences Designated as Future**

| Sentences | Classification |
|---|---|
| And so, my budget sets aside almost a trillion dollars over 10 years for additional needs. | FUTURE |
| We must set tough, world-class academic and occupational standards for all our children and give our teachers and students the tools they need to meet them." | FUTURE |

| And I propose we make a major investment in conservation by fully funding the Land and Water Conservation Fund. | FUTURE |
|---|---|

**A.14**: **Near Future vs. Distant Future Annotation Policy**
An important point is that our future annotation policy did not differentiate between the near and distant future. This can be seen in regard to how imperative sentences were classified. At its core, the imperative refers to direct commands or requests that are directed towards a goal or object (Alcázar and Saltarelli 2014; Bybee, Perkins, and Pagliuca 1994:179)**.** They can occur within simple sentences, where the subject is stated or implied, or among more complex sentences. The following examples illustrate this annotation policy

**Figure A1: Model Imperative Sentences for Classifying Near Future**

dobj
det
dative
amod
compound
Bring VERB
me PRON
a DET
14th- ADJ
century NOUN
clock. NOUN

In the example above, the subject is implied and is anchored on the base verb: (bring). A command is being issued that desires something that emanates from the speaker towards a future point that is not currently present. If the above type of sentence did not have a temporal present adjunct (e.g. 'now') in the sentence, then such sentence types were annotated as future. This exemplifies what we mean, when we say that we don't differentiate between the near and distant future. Below are additional examples of sentences used to illustrate this policy.

**Table A5: Model Imperative Sentences Classified as Future**

| Sentences | Model Classification |
|---|---|
| Let us agree to bridge old divides. | FUTURE |
| Let's do it together. | FUTURE |
| Let us, by our example, teach them to obey the law, respect our neighbors, and cherish our values. | FUTURE |

**A.15 Annotation Policy for Present References**
The present is a nebulous entity. Outside of linguistics, social psychologists, such as George Herbert Mead argued that the present is not a "knife edge", but rather composed of the imagined past and future (Mead 1932). In syntax, this admixture can occur when you try to differentiate present sentences from sentences that reference the recent past or near future. In the exemplars used, we addressed this issue by emphasizing that present perfect sentences should not be given a present classification, but rather placed in the past category. However, present-continuous sentences should be placed in the present, as opposed to the future.

**Table A6: Model Present Sentences Classified as Past or Present**

| Sentence | Tense/Aspect/ Stative | Model Classification |
|---|---|---|
| I am eating breakfast. | Present Continuous | Present |
| I have eaten breakfast. | Present Perfect | Past |
| And that's the whole idea. | Stative | Present |
| To make sure the retirement savings of America's seniors are not diverted in any other program, my budget protects all $2.6 trillion of the Social Security surplus for Social Security, and for Social Security alone. | Stative | Present |
| Yet the cause of freedom rests on more than our ability to defend ourselves and our allies. | Stative | Present |

**A.16 Classification for Conflicting Temporal References**

This issue was first addressed by focusing on the principle that phenomenologists have long defined as *intentionality* (Husserl 2014:62; Searle 1989). Intentionality refers to a directed form of action—one that begins from an egocentric point and projects toward a specific end. In the context of language structure, our focus is on identifying the speaker's dominant intention and the corresponding temporal anchor point or frame that underlies this domination. In other words, what is the speaker's primary-ultimate temporal goal (where is the end point of their dominant eventive goal). For example, the sentence below from our dataset served as a model used in our annotation for this process:

> *After years of leaders whose rhetoric attacked bureaucracy but whose action expanded it, we will actually reduce it by 252,000 people over the next five years.*

In the above, the speaker is referencing the past twice in order to instill the reader with a precedent that is being used for his or her primary goal, which is "we will reduce it by 250,000 people over the next five years." From a temporal point of view, the speaker is interested primarily in something that is going to happen in the future. Stated another way, the past is subordinate in importance to the future. Therefore, the primary classification of the sentence was given a future designation.

## A.17 Past Sentence Generation Template and Generated Examples

**Table A7: Past-Sentence Template Generator**

(8)

**Pa_template**

'{} before we lost the {}, we believed that if we {} {} enough, we {} have been successful.'

(9)

**pa_adjuncts**

['Yesterday', 'Last year', 'Last month', 'Last week', 'In a previous era', 'When we were inexperienced']

(10)

**pa_modals** = ['could', 'should']

**Table A8: Past Oriented Sentence Generated from Template**

| **Temporal Adverbial** | **Adverbial Time Clause** | **Subordinate Clause** | **Main Clause** |
|---|---|---|---|
| Last year | before we lost the election | we believed that if we struggled hard enough | we might have been successful |
| Last year | before we lost the election | we believed that if we campaigned hard enough | we could have been successful |
| Yesterday | before we lost the contract | we believed that if we struggled hard enough | we could have been successful. |
| Yesterday | before we lost the contract | we believed that if we struggled hard enough | we might have been successful |

### A.18 Present Sentence Generation Template and Generated Examples

**Table A9: Present-Sentence Template Generator**

4.

```
pres_vbs = ['walking', 'buying', 'traveling', 'eating', 'studying', 'wand
ering', 'defending', 'building', 'campaigning', 'dancing', 'singing', 'fl
ying', 'driving', 'skiing', 'boating']
```

5.

```
pres_adverbs = ['infrequently', 'slowly', 'occasionally', 'intensely', 'r
apidly', 'repeatedly', 'intermittently', 'frequently', 'habitually']
```

6.

```
pres_adjuncts = ['now', 'today', 'at the moment', 'at present']
```

**Table A10: Present Sentence Generated Examples**

| **Subject** | **Present Verb** | **Present Adverb** | **Present Adjunct** |
| --- | --- | --- | --- |
| We | are walking | infrequently | at the moment. |
| We | are walking | infrequently | at the present. |
| We | are walking | slow | now. |
| We | are walking | slowly | Today. |

### A.19 Complete Numerical SHAP Values.

**Table A14: Numerical SHAP Values**

| Rank | Feature | Importance |
| --- | --- | --- |
| 1 | VBD | 1.664185 |
| 2 | VB | 0.748659 |
| 3 | MD | 0.503552 |
| 4 | VBP | 0.379765 |
| 5 | VBN | 0.378118 |
| 6 | punct | 0.255277 |
| 7 | amod | 0.211524 |
| 8 | VBZ | 0.200538 |
| 9 | fu_3 | 0.1822 |
| 10 | NNP | 0.162308 |
| 11 | PRP$ | 0.159578 |
| 12 | PRP | 0.156923 |
| 13 | aux | 0.155046 |
| 14 | TO | 0.153113 |
| 15 | advcl | 0.123179 |
| 16 | advmod | 0.11803 |
| 17 | JJ | 0.099034 |
| 18 | NN | 0.097591 |
| 19 | ccomp | 0.093973 |
| 20 | compound | 0.093768 |
| 21 | DT | 0.093238 |
| 22 | conj | 0.092936 |
| 23 | NNS | 0.089024 |
| 24 | nsubj | 0.084782 |
| 25 | fu_5 | 0.082198 |
| 26 | xcomp | 0.077193 |
| 27 | dobj | 0.060078 |
| 28 | CC | 0.059756 |
| 29 | WP | 0.058551 |
| 30 | VBG | 0.057387 |
| 31 | neg | 0.052272 |
| 32 | NNPS | 0.051486 |

| 33 | poss | 0.051131 |
|---|---|---|
| 34 | prep | 0.05087 |
| 35 | relcl | 0.046928 |
| 36 | IN | 0.04692 |
| 37 | RB | 0.046329 |
| 38 | fu_4 | 0.045309 |
| 39 | pobj | 0.045 |
| 40 | attr | 0.043043 |